\documentclass[conference]{IEEEtran}
\IEEEoverridecommandlockouts
\usepackage{cite}
\usepackage[utf8]{inputenc}

\usepackage{graphicx}              
\usepackage{array}                 
\usepackage{booktabs}              
\usepackage{caption}               
\usepackage{amsmath,amssymb,amsfonts}
\usepackage{algorithmic}
\usepackage{float,graphicx}
\usepackage{textcomp}
\usepackage{hyperref}
\usepackage{xcolor}
\usepackage{booktabs}
\usepackage{multirow}
\def\BibTeX{{\rm B\kern-.05em{\sc i\kern-.025em b}\kern-.08em
    T\kern-.1667em\lower.7ex\hbox{E}\kern-.125emX}}
\usepackage{listings}
\usepackage{xcolor}
\usepackage{fancyhdr}
\begin{document}

\title{Enhancing Diffusion Models for High-Quality Image Generation}

\author{
    Jaineet Shah\quad Michael Gromis\quad Rickston Pinto 
}

\maketitle
\begin{abstract}
\textit{
This report presents the comprehensive implementation, evaluation, and optimization of Denoising Diffusion Probabilistic Models (DDPMs) and Denoising Diffusion Implicit Models (DDIMs), which are state-of-the-art generative models. During inference, these models take random noise as input and iteratively generate high-quality images as output. The study focuses on enhancing their generative capabilities by incorporating advanced techniques such as Classifier-Free Guidance (CFG), Latent Diffusion Models with Variational Autoencoders (VAE), and alternative noise scheduling strategies. The motivation behind this work is the growing demand for efficient and scalable generative AI models that can produce realistic images across diverse datasets, addressing challenges in applications such as art creation, image synthesis, and data augmentation. Evaluations were conducted on datasets including CIFAR-10 and ImageNet-100, with a focus on improving inference speed, computational efficiency, and image quality metrics like Fréchet Inception Distance (FID). Results demonstrate that DDIM + CFG achieves faster inference and superior image quality. Challenges with VAE and noise scheduling are also highlighted, suggesting opportunities for future optimization. This work lays the groundwork for developing scalable, efficient, and high-quality generative AI systems to benefit industries ranging from entertainment to robotics.
}\end{abstract}

\section{Introduction}
Diffusion models are a class of generative models that have revolutionized the "Generative AI" field by enabling high-quality data generation across various domains. These models operate by introducing random noise to datasets and subsequently learning to reverse this process to remove noise iteratively. With their noise prediction capability, diffusion models are capable of reconstructing original data from noisy inputs, making them versatile tools in image synthesis, data restoration, media content generation, and even robotic manipulation.

The primary goal of this project is to implement and enhance Denoising Diffusion Probabilistic Models (DDPMs) for high-quality image generation. During the training process, the inputs to these models consist of an RGB image per trading instance and Gaussian noise, progressively added to simulate the diffusion process. The model learns to predict the noise added to the image at each step, effectively learning how to reverse the noise-adding process and reconstruct the original image. During inference, the model starts with a random Gaussian noise input and iteratively denoises it step by step, guided by the learned noise predictions. This reverse process generates a high-quality image, starting from pure noise and refining it into a visually coherent output. Our implementation focuses on addressing computational inefficiencies inherent in diffusion models and optimizing their performance for practical deployment. This is achieved through integrating advanced methodologies, such as Denoising Diffusion Implicit Models (DDIMs), Latent Space Models/Variational Autoencoders (VAEs), and Classifier-Free Guidance (CFG), as outlined in the HW5 write-up.

Furthermore, this project introduces an exploratory investigation into modifying the linear noise scheduler of DDPMs. By optimizing the noise scheduling process for specific tasks or datasets, the study aims to uncover potential improvements in inference speed, computational efficiency, and robustness. These enhancements are critical for real-world applications, such as scalable and controllable image generation in industries like gaming, design, and advertising, where performance and quality are paramount.

This work not only reinforces the foundational understanding of diffusion models but also contributes to the ongoing research in improving their scalability and applicability for diverse real-world use cases.

\section{Literature Review}

Several key works have laid the foundation and advanced this field significantly. The introduction of denoising diffusion probabilistic models (DDPMs) by Ho et al. \cite{ho2020denoising} marked a breakthrough in generative modeling. By framing the generation process as the reverse of a noise corruption process, DDPMs demonstrated the ability to generate high-quality samples from complex distributions. This work highlighted the importance of optimizing the variational lower bound for improved training stability and image quality.

Expanding upon this framework, Song et al. \cite{song2020denoising} proposed denoising diffusion implicit models (DDIMs), which introduced a deterministic sampling process. This approach retained the high fidelity of DDPMs while significantly reducing the number of sampling steps. Notably, DDIMs have been shown to achieve comparable results to DDPMs with improved computational efficiency, making them more practical for large-scale applications.

To address the computational challenges of pixel-space diffusion, Rombach et al. \cite{rombach2022high} introduced latent diffusion models (LDMs). This approach integrates diffusion processes into a compressed latent space, significantly reducing the memory and computational requirements while maintaining image fidelity. LDMs have enabled high-resolution image synthesis and are particularly effective in tasks requiring fine-grained control, such as text-to-image generation.

The concept of leveraging latent representations also finds strong parallels with the foundational work of Kingma et al. \cite{kingma2013auto}, which introduced the Variational Autoencoder (VAE) framework. VAEs use a probabilistic latent space to model data distributions, effectively encoding complex structures into compressed representations. This idea underpins many advancements in generative modeling, including the latent space optimization techniques seen in LDMs \cite{rombach2022high}. By combining latent representations with diffusion processes, LDMs extend the utility of VAEs to handle more intricate generative tasks, such as high-resolution image synthesis.

Noise scheduling plays a critical role in the performance of diffusion models. The cosine noise scheduler, introduced by Nichol and Dhariwal \cite{nichol2021improved}, is a significant improvement over the original linear scheduling method proposed by Ho et al. \cite{ho2020denoising}. This method uses a cosine function to control the variance schedule, enabling smoother transitions between noise levels and enhancing sample quality. The cosine scheduler has become a standard technique in state-of-the-art diffusion models, contributing to better performance with minimal computational overhead.

In our project, we are implementing these papers to build a robust image generation pipeline. Leveraging the foundational principles of DDPMs \cite{ho2020denoising} and DDIMs \cite{song2020denoising}, we aim to optimize our generative models for both quality and efficiency. By integrating the cosine noise scheduling strategy from Nichol and Dhariwal \cite{nichol2021improved}, we are focusing on achieving smoother noise transitions and improved sample quality. Additionally, the latent space optimization introduced in LDMs \cite{rombach2022high} provide promising avenues for handling high-resolution synthesis tasks efficiently.

Our objective is to reproduce the high-quality results demonstrated in these papers while tailoring the approaches to our specific dataset and application needs. This involves experimenting with different noise scheduling strategies, sampling methods, and architectural choices to fine-tune the model for optimal performance. These state-of-the-art techniques form the backbone of our methodology, and by systematically implementing and testing them, we hope to achieve results that align with the high standards set by the research community.

\section{Dataset}

\subsection{Overview}

Our study utilizes two datasets: ImageNet-100 and CIFAR-10, to train various  models which will be discussed later in this paper. The lighterweight CIFAR-10 model was used to train less efficient models, while a more efficient model was trained on the heavier ImageNet-100.

\subsubsection{ImageNet-100}
ImageNet-100 is a subset of the larger ImageNet dataset, consisting of approximately 130,000 samples across 100 classes. Images are scaled to a resolution of 128 × 128. This dataset provides a wide range of categories, including animals, objects, and scenes, making it ideal for assessing the generalization capabilities of the model. ImageNet-100 was primarily used to train the final model and to evaluate its quality and robustness.

\begin{figure}[h!]
    \centering
    \includegraphics[width=1.0\columnwidth]{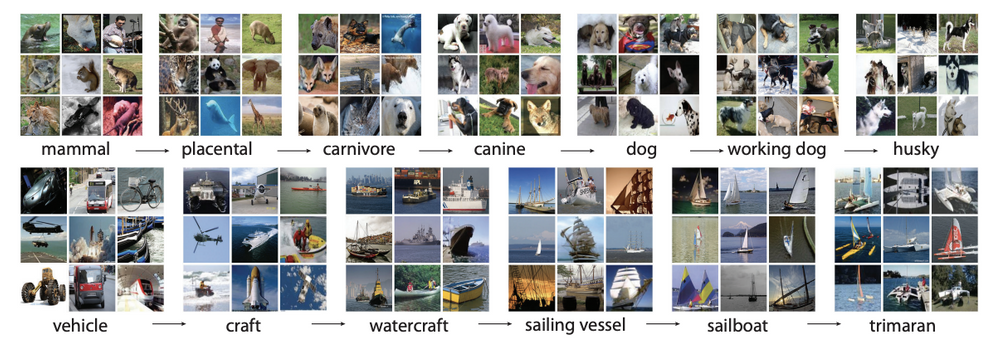} 
    \caption{Example images and classes from the ImageNet-100 dataset \cite{imagenet_cvpr09}}
    \label{fig:imagenet100}
\end{figure}

\subsubsection{CIFAR-10}
In the intermediate stages of model development, the CIFAR-10 dataset was employed. CIFAR-10 is a lightweight dataset containing 60,000 32 × 32 color images evenly distributed across 10 balanced classes. Its simplicity and smaller scale made it suitable for training intermediate versions of the model before transitioning to ImageNet-100 for final evaluation and performance optimization.

\begin{figure}[h!]
    \centering
    \includegraphics[width=0.9\columnwidth]{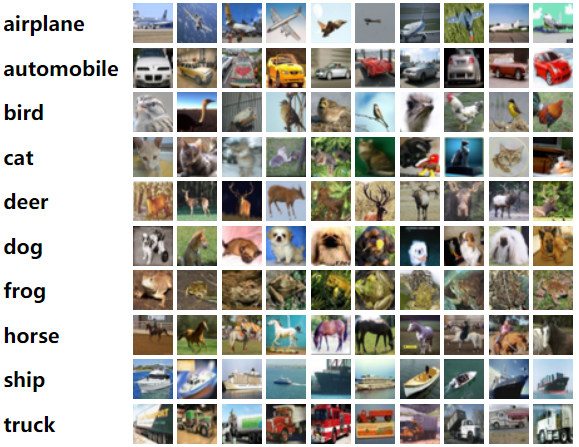} 
    \caption{Example images from the CIFAR-10 dataset \cite{cifar10_krizhevsky}. The smaller image size and reduced class diversity made it ideal for intermediate training stages.}
    \label{fig:cifar10}
\end{figure}

\subsection{Preprocessing and Collation}

The two datasets were easy to download and unzip and begin working with. For preprocessing, we resized the images to 128x128 pixels. No padding was applied, as the images within each dataset are uniform in size. We used minor data augmentation techniques, such as random horizontal flipping, and normalized the data to have zero mean and unit variance. Future work could explore additional augmentation techniques to improve model robustness. 

We adopt mini-batch sampling during training, a common practice in deep learning. We use a batch size of 128. The collation process involves shuffling the dataset at the start of each epoch to ensure randomness and prevent the model from learning patterns based on the data order.

\section{DDPM - Baseline Model}

\subsection{Baseline Model Selection }
We have decided to use our DDPM model as the baseline model. This choice is based on the fact that all the enhancements we described incorporate DDPM as a common architecture, making it a convenient and consistent reference point.

Denoising Diffusion Probabilistic Models (DDPMs) generate images by gradually reversing a diffusion process that corrupts real data with Gaussian noise over multiple timesteps. During training, the model learns to predict and denoise the added noise, and during sampling, it starts from pure noise and iteratively reconstructs realistic images by reversing the learned process.

\subsection{Evaluation Metrics: IS and FID in Diffusion Models}

To measure model performance, we use Inception Score (IS) and Fréchet Inception Distance (FID). These metrics are commonly used in the literature to evaluate generative models, particularly for assessing the quality and diversity of generated images.

\subsubsection{Inception Score (IS)}
The Inception Score (IS) is a widely used metric for evaluating the quality and diversity of generated images. It utilizes a pre-trained Inception network to compute the class probabilities of the generated images. The formula for IS is:

\[
\text{IS} = \exp\left( \mathbb{E}_{x \sim p_g(x)} \left[ D_{\text{KL}}\left(p(y|x) \,||\, p(y)\right) \right] \right),
\]

where:
\[
p(y|x) \text{ is the conditional class distribution for an image } x,
\]
\[
p(y) = \mathbb{E}_{x \sim p_g(x)}[p(y|x)] \text{ is the marginal class distribution,}
\]
and \(D_{\text{KL}}\) denotes the Kullback-Leibler divergence.

\subsubsection{Fréchet Inception Distance (FID)}
The Fréchet Inception Distance (FID) is a robust metric for evaluating the similarity between real and generated image distributions. It compares the feature distributions of real and generated images extracted using a pre-trained Inception network. The formula for FID is:

\[
\text{FID} = ||\mu_r - \mu_g||_2^2 + \text{Tr}(\Sigma_r + \Sigma_g - 2(\Sigma_r \Sigma_g)^{\frac{1}{2}}),
\]

where:
\[
\mu_r, \Sigma_r \text{ are the mean and covariance of real image features,}
\]
\[
\mu_g, \Sigma_g \text{ are the mean and covariance of generated image features}
\]

In the context of diffusion models, IS and FID are critical metrics for evaluating performance. Recall that Diffusion models aim to generate high-quality images that closely resemble real data. The Inception Score (IS) measures the confidence of class predictions for generated images, providing an assessment of image quality. Meanwhile, the Fréchet Inception Distance (FID) compares the distributions of real and generated images, offering valuable insight into both diversity and fidelity. As standard metrics in the literature, IS and FID are widely used for benchmarking generative models, such as GANs, and for evaluating variations of diffusion models, including DDIM and LDM. However, both metrics have limitations: IS does not evaluate how well the generated data aligns with the real data distribution, while FID, though more robust, can be sensitive to the quality of the pre-trained Inception network used for feature extraction. By combining IS and FID, a comprehensive evaluation of diffusion models can be achieved, addressing both image quality and alignment with the real data distribution.

\subsection{Prior Baseline Performance}
 Relevant State-of-the-Art is summarized below.

\begin{table}[ht]
\centering
\scriptsize
\caption{Comparison of IS and FID Across Models on CIFAR-10 and ImageNet}
\label{tab:is_fid_comparison}
\begin{tabular}{|p{1.2cm}|p{1.2cm}|p{.9cm}|p{.9cm}|p{2cm}|}
\hline
\textbf{Model} & \textbf{Dataset} & \textbf{IS} & \textbf{FID} & \textbf{Source} \\ \hline

\textbf{DDPM}  & CIFAR-10         & 9.46        & 3.17         & Ho et al. (2020) \\ \hline

\textbf{DDIM (S=10)} & CIFAR-10    & -           & 13.36        & Song et al. (2021)         \\ \hline
\textbf{DDIM (S=50)} & CIFAR-10    & -           & 4.67         & Song et al. (2021)         \\ \hline
\textbf{DDIM (S=1000)} & CIFAR-10  & -           & 4.04         & Song et al. (2021)         \\ \hline

\textbf{LDM}   & ImageNet         & -           & 27.0         & Rombach et al. (2022)      \\ \hline

\end{tabular}
\end{table}

The table compares the performance of different diffusion models using the Inception Score (IS) and Fréchet Inception Distance (FID) across CIFAR-10 and ImageNet datasets. DDPM achieves an IS of 9.46 and an FID of 3.17 on CIFAR-10, showcasing high-quality and diverse image generation. DDIM results demonstrate improved efficiency at fewer sampling steps on CIFAR-10, with an FID of 13.36 at 10 steps, 4.67 at 50 steps, and 4.04 at 1000 steps, highlighting its ability to achieve high-quality results faster than DDPM. On ImageNet, LDM yields an FID of 27.0, reflecting strong performance on a higher-resolution and more complex dataset. This summary indicates that DDIM provides a good trade-off between speed and quality, while LDM shows competitive performance on large-scale image generation tasks.

\subsection{End-to-End Process}
Figure \ref{fig:ddpm_visualization} visualizes the end-to-end process for DDPM. This section will explain the following components of the pipeline: 

\begin{figure*}[t]
\centering
\includegraphics[width=0.8\linewidth]{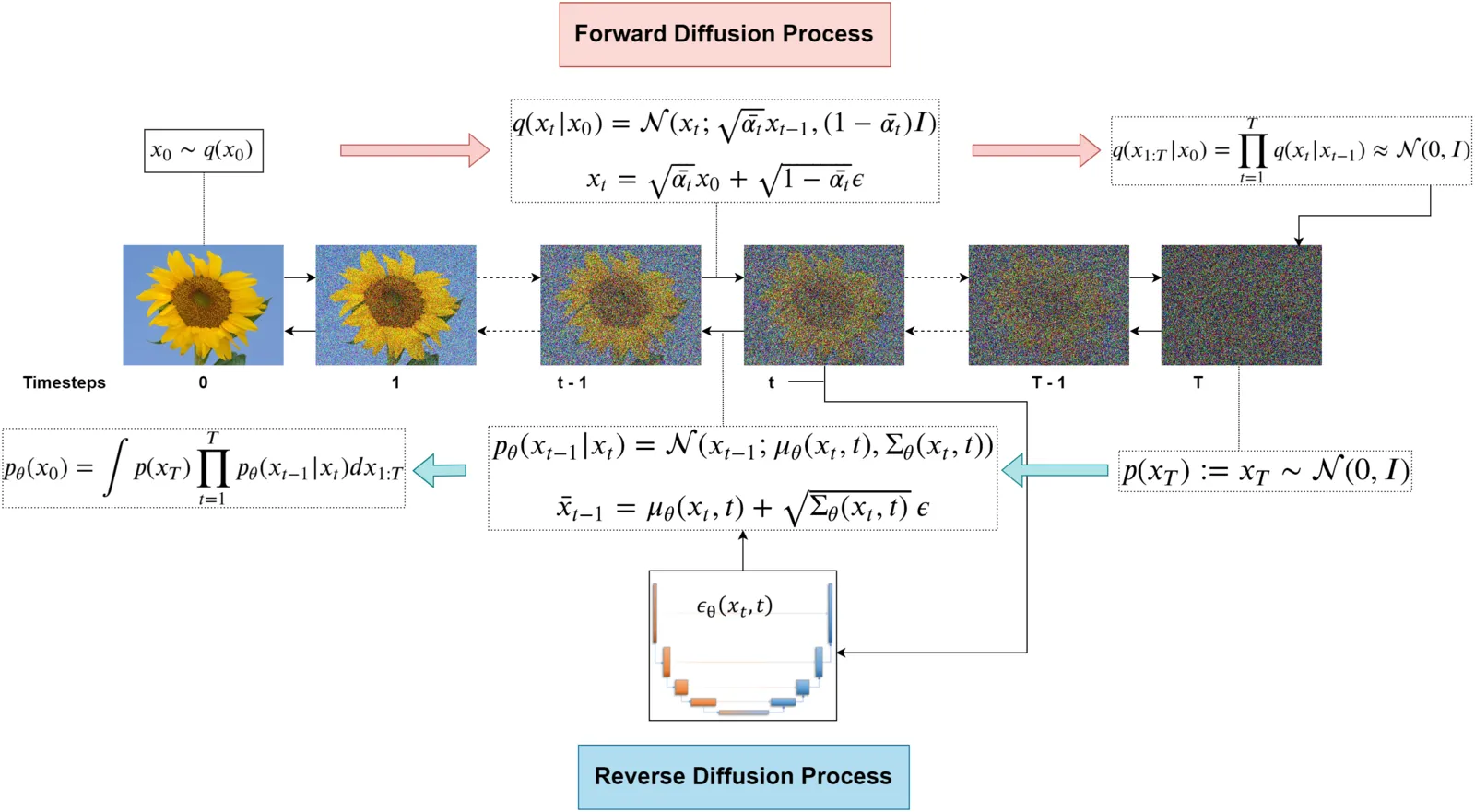} 
\caption{End-to-End DDPM Process Visualized, Source: \href{https://learnopencv.com/denoising-diffusion-probabilistic-models/}{LearnOpenCV}}
\label{fig:ddpm_visualization} 
\end{figure*}

\begin{enumerate}

    \item \textbf{Forward Diffusion Process:} Gradually adds Gaussian noise to the image over a series of timesteps, transforming the clean image into pure noise. This process is governed by the noise schedule (\(\beta_t\)), which determines the variance of the added noise at each step.

    \item \textbf{Reverse Diffusion Process:} Iteratively removes the noise added during the forward process to reconstruct clean data from noisy inputs. This reverse process is learned by the model, subsequently discussed, which predicts the noise at each timestep.

    \item \textbf{Noise Scheduler:} Controls the distribution of noise added during the forward process and its corresponding removal during the reverse process.

    \item \textbf{UNet Architecture:} Serves as the backbone of the model, taking noisy data and timestep information as input and predicting the noise to be removed. It features an encoder-decoder structure with Residual Connection blocks to preserve spatial details while processing multi-scale features.

    \item \textbf{Training Setup:} The model is trained to minimize the mean squared error (MSE) between the true noise and the noise predicted by the UNet. This allows the model to accurately denoise samples during the reverse process.

    \item \textbf{Inference Process:} Starts from pure Gaussian noise and iteratively applies the reverse diffusion process using the trained model to generate higher-quality images.

\end{enumerate}
\subsubsection{Forward Diffusion Process}

The forward diffusion process involves progressively corrupting the original data by adding Gaussian noise at each timestep \( t \). This process can be mathematically represented as:

\[
q(x_t | x_{t-1}) = \mathcal{N}(x_t; \sqrt{1 - \beta_t} x_{t-1}, \beta_t I),
\]

where:
\begin{itemize}
    \item \( x_t \) is the noisy sample at timestep \( t \),
    \item \( \beta_t \) is the variance of the noise added at timestep \( t \),
    \item \( \mathcal{N}(\cdot; \mu, \sigma^2) \) represents a Gaussian distribution with mean \( \mu \) and variance \( \sigma^2 \).
\end{itemize}

The cumulative effect of noise across timesteps is modeled as:

\[
q(x_t | x_0) = \mathcal{N}(x_t; \sqrt{\bar{\alpha}_t} x_0, (1 - \bar{\alpha}_t) I),
\]

where:
\begin{itemize}
    \item \(\bar{\alpha}_t = \prod_{i=1}^t (1 - \beta_i)\) is the cumulative product of \((1 - \beta)\), representing the proportion of the original signal retained at timestep \( t \).
\end{itemize}

The process for adding noise during the forward diffusion is outlined in Listing~\ref{lst:add_noise} and further explained here:

\begin{itemize}
    \item \textbf{Alpha Cumulative Product Calculation:}  
    The variable \(\texttt{sqrt\_alpha\_prod}\) is computed as the square root of the cumulative product of the alpha values:

    This represents the amount of signal retention (or "organization") at each timestep.

    \item \textbf{One Minus Alpha Cumulative Product Calculation:}  
    
    This term represents the amount of noise added at each timestep.

    \item \textbf{Noise Addition:}  
    The noisy samples are generated by a weighted sum of the original samples and the noise.

\end{itemize}

\begin{minipage}{\columnwidth}
\begin{lstlisting}[language=Python,label={lst:add_noise}, caption={Pseudocode for adding noise during the Forward Process},  basicstyle=\ttfamily\scriptsize, breaklines=true]
Function AddNoise(original_samples, noise, timesteps):
    # Compute square root of cumulative product of alpha at given timesteps
    sqrt_alpha_prod = sqrt(alphas_cumprod[timesteps])
    
    # Compute square root of (1 - cumulative product of alpha) at given timesteps
    sqrt_one_minus_alpha_prod = sqrt(1.0 - alphas_cumprod[timesteps])
    
    # Generate noisy samples by combining original samples and noise
    noisy_samples = (sqrt_alpha_prod * original_samples) + (sqrt_one_minus_alpha_prod * noise)
    
    Return noisy_samples
\end{lstlisting}
\end{minipage}

\subsubsection{Reverse Diffusion Process}

The reverse diffusion process is the core of the denoising mechanism in DDPMs. It reconstructs the original data by removing noise step-by-step, starting from pure noise. This process can be represented as:

\[
p_\theta(x_{t-1} | x_t) = \mathcal{N}(x_{t-1}; \mu_\theta(x_t, t), \Sigma_\theta(x_t, t)),
\]

where:
\begin{itemize}
    \item \(x_{t-1}\) is the reconstructed sample at timestep \(t-1\),
    \item \(\mu_\theta(x_t, t)\) is the predicted mean,
    \item \(\Sigma_\theta(x_t, t)\) is the predicted variance.
\end{itemize}

The predicted mean \(\mu_\theta(x_t, t)\) is computed as:
\[
\mu_\theta(x_t, t) = \frac{1}{\sqrt{\alpha_t}} \left(x_t - \frac{\beta_t}{\sqrt{1 - \bar{\alpha}_t}} \epsilon_\theta(x_t, t)\right),
\]
where:
\begin{itemize}
    \item \(\alpha_t = 1 - \beta_t\),
    \item \(\epsilon_\theta(x_t, t)\) is the predicted noise at timestep \( t \).
\end{itemize}

This iterative process continues until \(x_0\) (the clean sample) is reconstructed. We further explain our implementation in the inference portion of this discussion.

\subsubsection{Noise Scheduler}

The Noise Scheduler is a critical component of the forward and reverse diffusion process in DDPMs. Its primary purpose is to control how noise is added to the data during training by defining the variance \(\beta_t\) at each timestep \(t\). The scheduler ensures that noise is added progressively, starting with little noise at the early stages and increasing gradually in later timesteps. This noise schedule helps the model learn how to effectively reverse the noise process during inference. The noise scheduler used here is a linear noise schedule where the variance increases linearly from a small value \(\beta_\text{start}\) to a larger value \(\beta_\text{end}\). In the reverse process, the Noise Scheduler determines the scaling factors for each timestep, enabling the model to effectively remove noise in a structured manner, transitioning smoothly from coarse denoising to fine-grained reconstruction.The schedule can be expressed as:

\[
\beta_t = \beta_\text{start} + \left(\frac{t}{T} \right) (\beta_\text{end} - \beta_\text{start}),
\]

where:
\begin{itemize}
    \item \(t\) is the current timestep, ranging from 0 to \(T\),
    \item \(\beta_\text{start}\) and \(\beta_\text{end}\) are the starting and ending values of the noise variance.
\end{itemize}

This linear schedule ensures that noise gradually increases as the training progresses, giving the model sufficient time to denoise the data during reverse sampling.

Pseudocode for Linear Noise Scheduler:
The following pseudocode demonstrates how the linear schedule is implemented in the class `DDPMScheduler`:

\begin{minipage}{\linewidth}
\lstset{
    language=, 
    caption={Pseudocode for Linear Noise Scheduler}, 
    label={lst:linear_noise_scheduler},
    basicstyle=\ttfamily\scriptsize, 
    breaklines=true,
    frame=single,
    numbers=left,
    xleftmargin=2em,
    xrightmargin=2em,
    keywordstyle=\color{blue},
    commentstyle=\color{green!50!black},
    stringstyle=\color{red},
}
\begin{lstlisting}
Class DDPMScheduler:
    Initialize(num_train_timesteps, beta_start, beta_end, beta_schedule):
        Set num_train_timesteps to total number of timesteps
        Set beta_start to the starting beta value
        Set beta_end to the ending beta value

        If beta_schedule == "linear":
            betas = LinearInterpolation(beta_start, beta_end, num_train_timesteps)

    Function LinearInterpolation(beta_start, beta_end, num_steps):
        Initialize betas as an empty list
        For i in range(0, num_steps):
            Compute beta = beta_start + (i / (num_steps - 1)) * (beta_end - beta_start)
            Append beta to betas
        Return betas
\end{lstlisting}
\end{minipage}

\subsubsection{U-Net Architecture} The core neural network backbone of the DDPM is built upon a U-Net architecture, which is characterized by:

\begin{itemize}
    \item \textbf{Encoder (Downsampling Path):} Extracts features while reducing spatial dimensions using layers of layers Res-Net blocks and downsampling operations. Captures both detailed and broad patterns and stores intermediate feature maps for later use in the decoder.
    
    \item \textbf{Bottleneck:} Acts as a bridge between the encoder and decoder, processing the most compressed representation of the data. It captures global context and complex relationships.
    \item \textbf{Decoder (Upsampling Path):} Reconstructs the input to its original size by combining upsampled features with stored encoder feature maps through Res-Net blocks, promoting a balance of fine details and contextual coherence in the output.
\end{itemize}

Our UNet architecture, which takes an input tensor of shape \((N, C, H, W)\) where \(N\) is the batch size, \(C\) is the number of channels, and \(H, W\) are the height and width of the input image, is shown in Table \ref{tab:unet_architecture}. The model is well sized, containing 35M total parameters.

\begin{table*}[htbp!]
\centering
\scriptsize 
\caption{UNet Model Architecture and Parameters for DDPM - CIFAR-10}
\label{tab:unet_architecture}
\begin{tabular}{|p{3cm}|p{6cm}|p{6cm}|}
\hline
\textbf{Model}                & \textbf{Hidden Layers}                                                         & \textbf{Parameters / Hyperparameters}                                  \\ \hline
Embedding Layer               & TimeEmbedding and Conv2D                                                      & -                                                                     \\ \hline
Encoder                   & 3 x ResBlock (64), 1 x DownSample, 1 x ResBlock (64), 2 x ResBlock (128), 1 x DownSample, 1 x ResBlock (128), 2 x ResBlock (256), 1 x DownSample, 3 x ResBlock (256) & Number of kernels: 64, 128, and 256                                   \\ \hline
Bottleneck                    & 2 x ResBlock (256)                                                            & Number of kernels: 256                                                \\ \hline
Decoder                    & 4 x ResBlock (512), 1 x UpSample, 3 x ResBlock (512), 1 x ResBlock (384), 1 x UpSample, 1 x ResBlock (384), 2 x ResBlock (256), 1 x ResBlock (192), 1 x UpSample, 1 x ResBlock (192), 3 x ResBlock (128) & Number of kernels: 128, 192, 384, and 512                             \\ \hline
FinalLayer                    & 1 x GroupNorm, 1 x SiLU, 1 x Conv2D (64)                                      & Number of kernels: 64                                                 \\ \hline
\end{tabular}
\end{table*}

\subsubsection{Training Setup}
In this section, we outline the training process, beginning with our objective function.

We use a formulation of Mean Squared Error (MSE) for our loss function during DDPM training:

\[
\mathcal{L}_{\text{simple}}(\theta) := \mathbb{E}_{t, x_0, \epsilon} \left[ \left\| \epsilon - \epsilon_\theta \left( \sqrt{\bar{\alpha}_t} x_0 + \sqrt{1 - \bar{\alpha}_t} \epsilon, t \right) \right\|^2 \right]
\]

\subsection*{Description of Components}

\begin{itemize}
    \item \textbf{\(\mathcal{L}_{\text{simple}}(\theta)\):}
    The  loss function that minimizes the discrepancy between the true noise \(\epsilon\) and the model's predicted noise \(\epsilon_\theta\).

    \item \textbf{\(\mathbb{E}_{t, x_0, \epsilon}\):}
    The expectation is taken over:
    \begin{itemize}
        \item \(t\): A timestep uniformly sampled from \(\{1, \dots, T\}\), where \(T\) is the total number of timesteps.
        \item \(x_0\): The original data sample drawn from the training dataset.
        \item \(\epsilon\): The Gaussian noise sampled from a standard normal distribution, \(\epsilon \sim \mathcal{N}(0, I)\).
    \end{itemize}

    \item \textbf{\(\epsilon\):}
    The Gaussian noise added to the original data \(x_0\) during the forward diffusion process.

    \item \textbf{\(\epsilon_\theta(x_t, t)\):}
    The model’s prediction of the noise \(\epsilon\) at timestep \(t\), conditioned on the noisy data \(x_t\). The parameters \(\theta\) are optimized during training.

    \item \textbf{\(\sqrt{\bar{\alpha}_t} x_0 + \sqrt{1 - \bar{\alpha}_t} \epsilon\):}
    The noisy data \(x_t\) at timestep \(t\), constructed as a mixture of:
    \begin{itemize}
        \item \(\sqrt{\bar{\alpha}_t} x_0\): A scaled version of the original data \(x_0\).
        \item \(\sqrt{1 - \bar{\alpha}_t} \epsilon\): A scaled version of the noise \(\epsilon\).
    \end{itemize}

\end{itemize}

By minimizing this loss, the model can learn to reverse the forward diffusion process step-by-step, ultimately reconstructing high-quality data from pure noise during inference. Pseudocode below describes the training process:

\begin{minipage}{\linewidth}
\lstset{
    language=, 
    caption={Pseudocode for Training DDPM}, 
    label={lst:train_ddpm},
    basicstyle=\ttfamily\scriptsize, 
    breaklines=true,
    frame=single,
    numbers=left,
    xleftmargin=2em,
    xrightmargin=2em,
    keywordstyle=\color{blue},
    commentstyle=\color{green!50!black},
    stringstyle=\color{red},
}
\begin{lstlisting}
Function TrainDDPM(model, data_loader, num_epochs, optimizer, loss_function, noise_scheduler):
    For each epoch in num_epochs:
        Print("Starting epoch {epoch}...")

        For each batch in data_loader:
            # Step 1: Get real images from the batch
            real_images = batch.images 

            # Step 2: Sample random timesteps
            timesteps = RandomUniform(0, T, batch_size)

            # Step 3: Generate random noise
            noise = RandomNormal(0, 1, shape_of(real_images))

            # Step 4: Add noise to real images
            noisy_images = noise_scheduler.add_noise(real_images, noise, timesteps)

            # Step 5: Predict the noise using the model
            predicted_noise = model.predict(noisy_images, timesteps)

            # Step 6: Calculate loss between predicted noise and true noise
            loss = loss_function(predicted_noise, noise)

            # Step 7: Update model weights
            optimizer.zero_grad()
            loss.backward()
            optimizer.step()

        Print("Epoch {epoch} complete with loss {loss}.")
    Return model
\end{lstlisting}
\end{minipage}

In summary, for each batch during training, the real images are first corrupted by adding Gaussian noise at randomly sampled timesteps according to the noise scheduler. The model then predicts the noise added to the images, and the loss function measures the discrepancy between the predicted and true noise. Finally, the model's parameters are updated via backpropagation to minimize this loss, enabling it to progressively learn the reverse diffusion process.

Initial training shows that training the DDPM model on ImageNet-100 would take about a week to complete (2000 epochs), so we decided to train it on the CIFAR-10 dataset which not only has fewer images, but also smaller resolution. This dataset is consistent with what the original authors used to train their model. Our hyperparameters are summarized in Table~\ref{tab:hyperparameters}.

\begin{table}[ht]
\centering
\scriptsize
\caption{Hyperparameters for Training and Model Configuration}
\label{tab:hyperparameters}
\begin{tabular}{|l|l|l|}
\hline
\textbf{Hyperparameter}    & \textbf{Value}         & \textbf{Description}            \\ \hline
\texttt{image\_size}       & 32                    & Image size                      \\ \hline
\texttt{batch\_size}       & 128                   & Batch size                      \\ \hline
\texttt{num\_workers}      & 4                     & Number of data loading workers  \\ \hline
\texttt{num\_classes}      & 10                    & Number of output classes        \\ \hline
\texttt{num\_epochs}       & 480                   & Number of training epochs       \\ \hline
\texttt{learning\_rate}    & $1 \times 10^{-4}$    & Learning rate                   \\ \hline
\texttt{weight\_decay}     & $1 \times 10^{-4}$    & Weight decay                    \\ \hline
\texttt{num\_train\_timesteps} & 1000             & Number of training timesteps    \\ \hline
\texttt{num\_inference\_steps} & 250              & Number of inference timesteps   \\ \hline
\texttt{beta\_start}       & 0.0002               & Starting beta value             \\ \hline
\texttt{beta\_end}         & 0.02                 & Ending beta value               \\ \hline
\texttt{beta\_schedule}    & \texttt{linear}      & Beta schedule                   \\ \hline
\texttt{variance\_type}    & \texttt{fixed\_small}& Variance type                   \\ \hline
\texttt{predictor\_type}   & \texttt{epsilon}     & Predictor type                  \\ \hline
\texttt{unet\_in\_size}    & 32                   & Input size for UNet             \\ \hline
\texttt{unet\_in\_ch}      & 3                    & Number of input channels        \\ \hline
\texttt{unet\_ch}          & 64                   & Base number of channels         \\ \hline
\texttt{unet\_num\_res\_blocks} & 3              & Number of residual blocks       \\ \hline
\texttt{unet\_ch\_mult}    & [1, 2, 4, 4]         & Channel multiplier for each level \\ \hline
\texttt{unet\_attn}        & [2, 3]               & Attention layers                \\ \hline
\texttt{unet\_dropout}     & 0.1                  & Dropout rate                    \\ \hline
\end{tabular}
\end{table}

\subsubsection{Inference Process}

The inference process in a Denoising Diffusion Probabilistic Model (DDPM) involves reversing the forward diffusion process to generate data from random noise. Below is a description of the key steps:

\begin{itemize}
    \item \textbf{Initialize with Random Noise:} Start with a randomly sampled noise tensor \( x_T \sim \mathcal{N}(0, I) \), representing the noisiest state at timestep \( T \).
    
    \item \textbf{Iterative Denoising:} For each timestep \( t \) (from \( T \) to 1):
    \begin{itemize}
        \item Use the trained model \( \epsilon_\theta(x_t, t) \) to predict the noise \( \epsilon \) present in \( x_t \).
        \item Estimate the intermediate sample \( x_{t-1} \) using:
        \[
        x_{t-1} = \frac{1}{\sqrt{\alpha_t}} \left( x_t - \frac{1 - \alpha_t}{\sqrt{1 - \bar{\alpha}_t}} \epsilon_\theta(x_t, t) \right) + \sigma_t z,
        \]
        where:
        \begin{itemize}
            \item \( \alpha_t \): Noise scaling factor.
            \item \( \bar{\alpha}_t \): Cumulative product of noise scalars.
            \item \( \sigma_t z \): Optional noise term added for stochasticity (only for variational inference).
        \end{itemize}
        \item For \( t > 1 \), Gaussian noise \( z \sim \mathcal{N}(0, I) \) is added to preserve variance. For \( t = 1 \), no noise is added to ensure a clean output.
    \end{itemize}
    
    \item \textbf{Generate Final Sample:} The process concludes at timestep \( t = 1 \), yielding \( x_0 \), which approximates a sample from the learned data distribution.

    This process is represented in pseudocode in pseudocode below:
\end{itemize}

\subsection*{Pseudocode for DDPM Inference}
\begin{minipage}{\linewidth}
\lstset{
    language=, 
    caption={Pseudocode for DDPM Inference Process}, 
    label={lst:ddpm_inference},
    basicstyle=\ttfamily\scriptsize, 
    breaklines=true,
    frame=single,
    numbers=left,
    xleftmargin=2em,
    xrightmargin=2em,
    keywordstyle=\color{blue},
    commentstyle=\color{green!50!black},
    stringstyle=\color{red},
}
\begin{lstlisting}
Function DDPMInference(model, noise_scheduler, T):
    # Step 1: Initialize with random noise
    x_T = RandomNormal(mean=0, std=1, shape)

    # Step 2: Iterative denoising from t = T to 1
    For t in range(T, 1, -1):
        # Predict noise in the sample
        predicted_noise = model.predict(x_t, t)

        # Compute the mean of the next step
        mean = (1 / sqrt(alpha_t)) * (x_t - ((1 - alpha_t) / sqrt(1 - alpha_bar_t)) * predicted_noise)
        
        # Add optional noise term for t > 1
        If t > 1:
            z = RandomNormal(mean=0, std=1, shape)
            x_{t-1} = mean + sigma_t * z
        Else:
            x_{t-1} = mean

    # Return the final generated sample x_0
    Return x_0
\end{lstlisting}
\end{minipage}

\section{Implemented Extensions}

\subsection{DDIM Implementation}
Building on DDPM, DDIM introduced deterministic reverse diffusion to enhance inference speed and consistency. Unlike DDPM, which relies on a Markov chain for reverse diffusion, DDIM introduces a deterministic path leveraging information from earlier timesteps, enabling intermediate step skipping.
Each timestep depends on the model's noise prediction and the deterministic reverse step formulation, avoiding randomness.

\subsubsection{Core Implementation Details}

Building upon the foundation established by DDPM, our implementation of DDIM (Deterministic Denoising Diffusion Implicit Models) introduces a deterministic reverse diffusion process. This modification transitions the sampling procedure from a probabilistically-driven, chaotic mechanism to a more controlled, deterministic process.

\begin{figure*}[ht]
\centering
\[
x_{t-1} = \sqrt{\alpha_{t-1}} 
\left(
\frac{x_t - \sqrt{1 - \alpha_t} \epsilon_\theta(x_t, t)}{\sqrt{\alpha_t}}
\right) 
+ \sqrt{1 - \alpha_{t-1} - \sigma_t^2} \cdot \epsilon_\theta(x_t, t) 
+ \sigma_t \cdot z_t,
\]
\caption{Reverse diffusion process for generating \(x_{t-1}\) from \(x_t\).}
\label{fig:reverse_diffusion}
\end{figure*}

The reverse diffusion process for generating \(x_{t-1}\) from \(x_t\) is defined in Figure~\ref{fig:reverse_diffusion}, where:
\begin{itemize}
    \item \(x_t\) is the noisy data at timestep \(t\).
    \item \(\epsilon_\theta(x_t, t)\) is the noise predicted by the model at timestep \(t\).
    \item \(\alpha_t\) is the cumulative noise schedule at timestep \(t\).
    \item \(\sigma_t\) controls the variance of the noise added at each step.
    \item \(z_t \sim \mathcal{N}(0, I)\) is sampled from a standard normal distribution, representing optional stochastic noise.
\end{itemize}

The variance for the DDIM reverse diffusion process also differs from DDPM and is defined as:

\[
\Sigma_t = \eta \cdot \frac{(1 - \bar{\alpha}_{t-1})}{(1 - \bar{\alpha}_t)} \cdot (1 - \alpha_t),
\]

where:
\begin{itemize}
    \item \(\eta\) is a hyperparameter controlling the degree of stochasticity in the reverse process. For deterministic DDIM sampling, \(\eta = 0\).
    \item \(\bar{\alpha}_t\) and \(\bar{\alpha}_{t-1}\) are the cumulative products of \(\alpha_t\) at timesteps \(t\) and \(t-1\), respectively.
    \item \(\alpha_t\) is the noise scaling factor at timestep \(t\).
\end{itemize}

This equation combines three key components:
\begin{enumerate}
    \item The predicted clean data, scaled by \(\sqrt{\alpha_{t-1}}\).
    \item Residual noise from the model prediction, scaled by \(\sqrt{1 - \alpha_{t-1} - \sigma_t^2}\).
    \item Additional Gaussian noise \(z_t\), scaled by \(\sigma_t\), for stochasticity.
\end{enumerate}

By iteratively applying this formula from \(T\) to 1, the reverse diffusion process reconstructs high-quality data from random noise.

As a result, this enables greater control and allows us to achieve higher-quality reconstructions while reducing the number of inference steps required.

The pseudocode in Listing~\ref{lst:ddim_scheduler_step} describes a single denoising step during the reverse diffusion process in DDIM. At each timestep, the function calculates the previous timestep, retrieves the cumulative noise and signal values, and predicts the clean sample using the noisy input and the model's predicted noise. This estimated sample is then returned to iteratively reconstruct the original data by progressively removing noise across timesteps similar to DDPM.

\begin{minipage}{\linewidth}
\lstset{
    language=, 
    caption={Pseudocode for DDIM modification in Scheduler Step Function}, 
    label={lst:ddim_scheduler_step},
    basicstyle=\ttfamily\scriptsize, 
    breaklines=true,
    frame=single,
    numbers=left,
    xleftmargin=2em,
    xrightmargin=2em,
    keywordstyle=\color{blue},
    commentstyle=\color{green!50!black},
    stringstyle=\color{red},
}
\begin{lstlisting}
Function Step(model_output, timestep, sample, eta=0.0):
    # Step 1: Calculate the prior timestep
    prev_t = previous_timestep(timestep)

    # Step 2: Compute the cumulative product of alphas
    alpha_prod_t = alphas_cumprod[timestep]
    alpha_prod_t_prev = alphas_cumprod[prev_t]
    beta_prod_t = 1 - alpha_prod_t

    # Step 3: Estimate the original sample based on the model's output
    pred_original_sample = (sample - sqrt(beta_prod_t) * model_output) / sqrt(alpha_prod_t)

    Return pred_original_sample
\end{lstlisting}
\end{minipage}

\subsection{Advanced Methodologies}
To further enhance diffusion models, the following were integrated:
\begin{itemize}
    \item \textbf{Latent Diffusion Models (VAE):} Reduced computational complexity by operating in latent space. Training leveraged a pre-trained VAE for ImageNet-100 that was provided to the team. The VAE works by taking an image and converting it into the latent space.
    \item \textbf{Classifier-Free Guidance (CFG):} Simplified conditional image generation by interpolating between conditional and unconditional scores.

With the VAE, we are finally able to train on ImageNet-100, as the VAE compresses the input of size 128x128 into a latent space of size 32×32. In Table~\ref{tab:unet_larger_kernels}, we present the model architecture used for the combined DDIM, VAE, and CFG model. Hyper parameters used for training this model are presented below.

\section*{Hyperparameter Table - Advanced Methods Model}

\begin{table}[h!]
\centering
\setlength{\tabcolsep}{4pt} 
\renewcommand{\arraystretch}{1.2} 
\begin{tabular}{| p{3cm} | p{2cm} | p{3.5cm} |}
\hline
\textbf{Hyperparameter} & \textbf{Value} & \textbf{Description} \\ \hline
\texttt{seed} & 42 & Random seed for reproducibility \\ \hline
\texttt{image\_size} & 128 & Image resolution \\ \hline
\texttt{batch\_size} & 128 & Number of samples per batch \\ \hline
\texttt{num\_workers} & 4 & Number of data loader workers \\ \hline
\texttt{num\_classes} & 101 & Number of output classes \\ \hline
\texttt{num\_epochs} & 51 & Total number of training epochs \\ \hline
\texttt{learning\_rate} & \texttt{1e-4} & Learning rate for optimizer \\ \hline
\texttt{weight\_decay} & \texttt{1e-4} & Weight decay (L2 regularization) \\ \hline
\texttt{num\_train\_timesteps} & 1000 & Number of timesteps for training \\ \hline
\texttt{num\_inference\_steps} & 250 & Number of timesteps for inference \\ \hline
\texttt{beta\_start} & 0.0002 & Initial beta value in the schedule \\ \hline
\texttt{beta\_end} & 0.02 & Final beta value in the schedule \\ \hline
\texttt{beta\_schedule} & \texttt{'linear'} & Beta schedule type \\ \hline
\texttt{variance\_type} & \texttt{fixed\_small} & Type of variance \\ \hline
\texttt{predictor\_type} & \texttt{epsilon} & Type of predictor \\ \hline
\texttt{unet\_in\_size} & 32 & Input size for the U-Net \\ \hline
\texttt{unet\_in\_ch} & 3 & Number of input channels for the U-Net \\ \hline
\texttt{unet\_ch} & 128 & Base channel count for the U-Net \\ \hline
\texttt{unet\_num\_res\_blocks} & 3 & Number of residual blocks per layer \\ \hline
\texttt{unet\_ch\_mult} & \texttt{[1, 2, 2, 4, 4]} & Channel multiplier for each layer \\ \hline
\texttt{unet\_attn} & \texttt{[1, 2, 3, 4]} & Layers with attention mechanisms \\ \hline
\texttt{unet\_dropout} & 0.1 & Dropout rate in the U-Net \\ \hline
\end{tabular}
\caption{Hyperparameters for the advanced methods configuration.}
\end{table}

\begin{table}[htbp!]
\centering
\scriptsize
\begin{tabular}{|p{2.5cm}|p{4cm}|p{3.5cm}|}
\hline
\textbf{Model}                & \textbf{Hidden Layers}                                                         & \textbf{Parameters / Hyperparameters}                                  \\ \hline
Embedding Layer               & TimeEmbedding and Conv2D                                                      & -                                                                     \\ \hline
Encoder                       & 3 x ResBlock (128), 1 x DownSample, 1 x ResBlock (128), 2 x ResBlock (256), 1 x DownSample, 3 x ResBlock (256), DownSample, 1 x ResBlock (256), 1 x DownSample, 2 x ResBlock (512), DownSample, 3 x ResBlock (512) & Number of kernels: 128, 256, and 512                                   \\ \hline
Bottleneck                    & 2 x ResBlock (512)                                                            & Number of kernels: 512                                                \\ \hline
Decoder                       & 4 x ResBlock (1024), 1 x UpSample, 3 x ResBlock (1024), 1 x ResBlock (768), 1 x UpSample, 1 x ResBlock (768), 3 x ResBlock (512), 1 x UpSample, 3 x ResBlock (512), 1 x ResBlock (368), 1 x UpSample, 1 x ResBlock (368), 3 x ResBlock (256)  & Number of kernels: 368, 512, 768, and 1024                              \\ \hline
FinalLayer                    & 1 x GroupNorm, 1 x SiLU, 1 x Conv2D (128)                                     & Number of kernels: 128                                               \\ \hline
\end{tabular}
\caption{UNet Model Architecture and Parameters for DDIM + VAE + CFG - ImageNet-100}
\label{tab:unet_larger_kernels}
\end{table}

\subsection{Exploration: Cosine Noise Scheduler}
For the exploration portion of this project, we implemented an alternative to DDPM's linear noise scheduler. Following the methodology described by Nichol and Dhariwal \cite{nichol2021improved}, we allocated noise based on a cosine function, placing different emphasis on timesteps throughout the denoising process. Specifically, the cosine scheduler assigns more noise to earlier timesteps, encouraging the model to focus on learning how to handle noisier, more degraded inputs effectively. Conversely, it reduces the noise added during later timesteps, allowing the model to refine and recover fine-grained details as the denoising progresses. This non-linear noise allocation aims to strike a balance between learning global structure early in the process and focusing on high-quality reconstruction in the final stages, leading to improved sample quality and more efficient training dynamics.

\subsection*{Definition of \(\bar{\alpha}_t\)}

In contrast to the linear schedule, the cumulative noise parameter \(\bar{\alpha}_t\) is defined using a cosine function:
\[
\bar{\alpha}_t = \cos^2\left(\frac{\pi}{2} \cdot \frac{t}{T}\right)
\]
where:
\begin{itemize}
    \item \(t\) is the current timestep,
    \item \(T\) is the total number of timesteps.
\end{itemize}

This ensures a smooth decay of \(\bar{\alpha}_t\) over time, leading to a more stable noise variance schedule.

\subsection*{Computation of \(\alpha_t\) and \(\beta_t\)}

From \(\bar{\alpha}_t\), the individual signal preservation parameter \(\alpha_t\) and noise variance parameter \(\beta_t\) are computed as follows:
\[
\alpha_t = \frac{\bar{\alpha}_t}{\bar{\alpha}_{t-1}}
\]
\[
\beta_t = 1 - \alpha_t
\]

Here:
\begin{itemize}
    \item \(\alpha_t\) represents the fraction of the signal preserved at timestep \(t\),
    \item \(\beta_t\) represents the noise variance added at timestep \(t\).
\end{itemize}

\subsection*{Adjusted Formula for Numerical Stability}

To avoid numerical instability during computation, a small constant \(\epsilon\) was added. This modifies the definition of \(\bar{\alpha}_t\) to:
\[
\bar{\alpha}_t = \frac{\cos^2\left(\frac{\pi}{2} \cdot \frac{t}{T}\right)}{\cos^2\left(\frac{\pi}{2} \cdot \frac{t - 1}{T}\right) + \epsilon}
\]
where \(\epsilon\) is a small positive constant added for numerical stability.

\end{itemize}

\section{Summary of Experiments}

\subsection{DDPM on CIFAR-10}
\begin{itemize}
    \item Trained a Denoising Diffusion Probabilistic Model (DDPM) on the CIFAR-10 dataset.
    \item Evaluated the model's performance in terms of Inception Score (IS) and Fréchet Inception Distance (FID).
    \item Served as the baseline diffusion approach for comparison with other methods.
\end{itemize}

\subsection{DDIM on CIFAR-10}
\begin{itemize}
    \item Utilized a Denoising Diffusion Implicit Model (DDIM) on CIFAR-10.
    \item Explored improved sampling efficiency compared to DDPM, leveraging non-Markovian steps.
    \item Analyzed how well DDIM maintains generation quality with faster sampling.
\end{itemize}

\subsection{DDIM + VAE + CFG on ImageNet}
\begin{itemize}
    \item Applied DDIM with pre-trained VAE and Classifier-Free Guidance (CFG) on ImageNet.
    \item Incorporated CFG to condition the generation process on class labels, enhancing fidelity and alignment with the desired class.
    \item Evaluated the ability of the model to generate high-resolution, class-consistent images.
\end{itemize}

\subsection{DDIM with Cosine Noise Schedule on CIFAR-10}
\begin{itemize}
    \item Implemented DDIM with a cosine noise schedule on CIFAR-10 to improve noise distribution during the diffusion process.
    \item Compared the performance of the cosine noise schedule against standard noise schedules.
    \item Measured improvements in sample quality and training stability.
\end{itemize}

\section{Results}

\subsection{Preliminary Results and Focused Experiments}

Given limited training time and resources, our models were able to converge for experiments B and C. For experiments A and D, we've only been able to train these models for about 10 epochs so far. Our results are too early to report, and so we are omitting our intermediate results. We urge our readers to focus on experiments B and C which successfully implement the diffusion process and show performance gains. We expand upon this in this section and the discussion.

\subsection{Experiment Tracking with Weights \& Biases}
All experimental runs, configurations, and metrics are tracked and logged using the Weights \& Biases platform. This ensures reproducibility and detailed analysis of training and evaluation processes.

The complete experiment logs can be accessed through the following link:

\begin{center}
    \href{https://wandb.ai/jaineets-carnegie-mellon-university/ddpm/table?nw=nwuserjaineets}{\textbf{Weights \& Biases Experiment Dashboard}}
\end{center}

This dashboard provides a comprehensive overview of the various ablations performed, loss curves of the models, detailed model configurations, and examples of generated output images.

\subsection{Quantitative Metrics}
Performance was assessed using FID and Inception Score (IS) across multiple configurations, summarized in Table~\ref{table:results}.

\begin{table}[H]
    \centering
    \renewcommand{\arraystretch}{1.2} 
    \setlength{\tabcolsep}{8pt} 
    \resizebox{\columnwidth}{!}{ 
    \begin{tabular}{|c|c|c|c|c|c|}
        \hline
        Dataset & Model & FID & IS & Steps & Batch Size \\
        \hline
        CIFAR-10 & DDIM (Baseline with Linear Noise Schedule) & 47.92 & 6.03 & 187,000 & 128 \\
        ImageNet-100 & DDIM + CFG + VAE & 323.51 & 2.51 & 52,000 & 128 \\
        \hline
    \end{tabular}
    }
    \caption{Performance metrics for various models.}
    \label{table:results}
\end{table}

\subsection{Qualitative Insights}
\begin{figure}[H]
    \centering
    \includegraphics[width=\columnwidth]{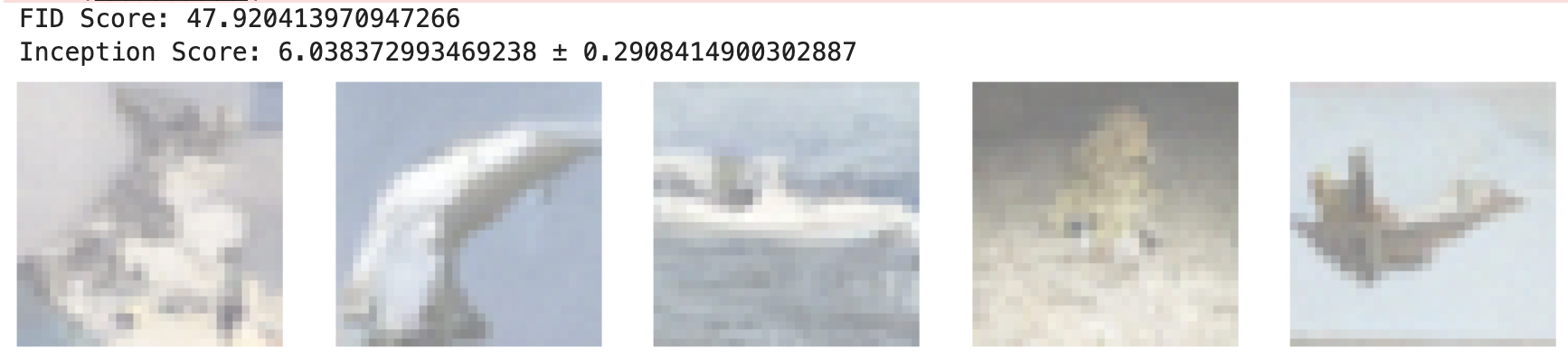} 
    \caption{Results of the DDIM model.}
    \label{fig:ddim_results}
\end{figure}

Qualitative results are presented in Figure~\ref{fig:ddim_results}, which shows examples of images generated by the DDIM model.

\begin{figure}[H]
    \centering
    \includegraphics[width=\columnwidth]{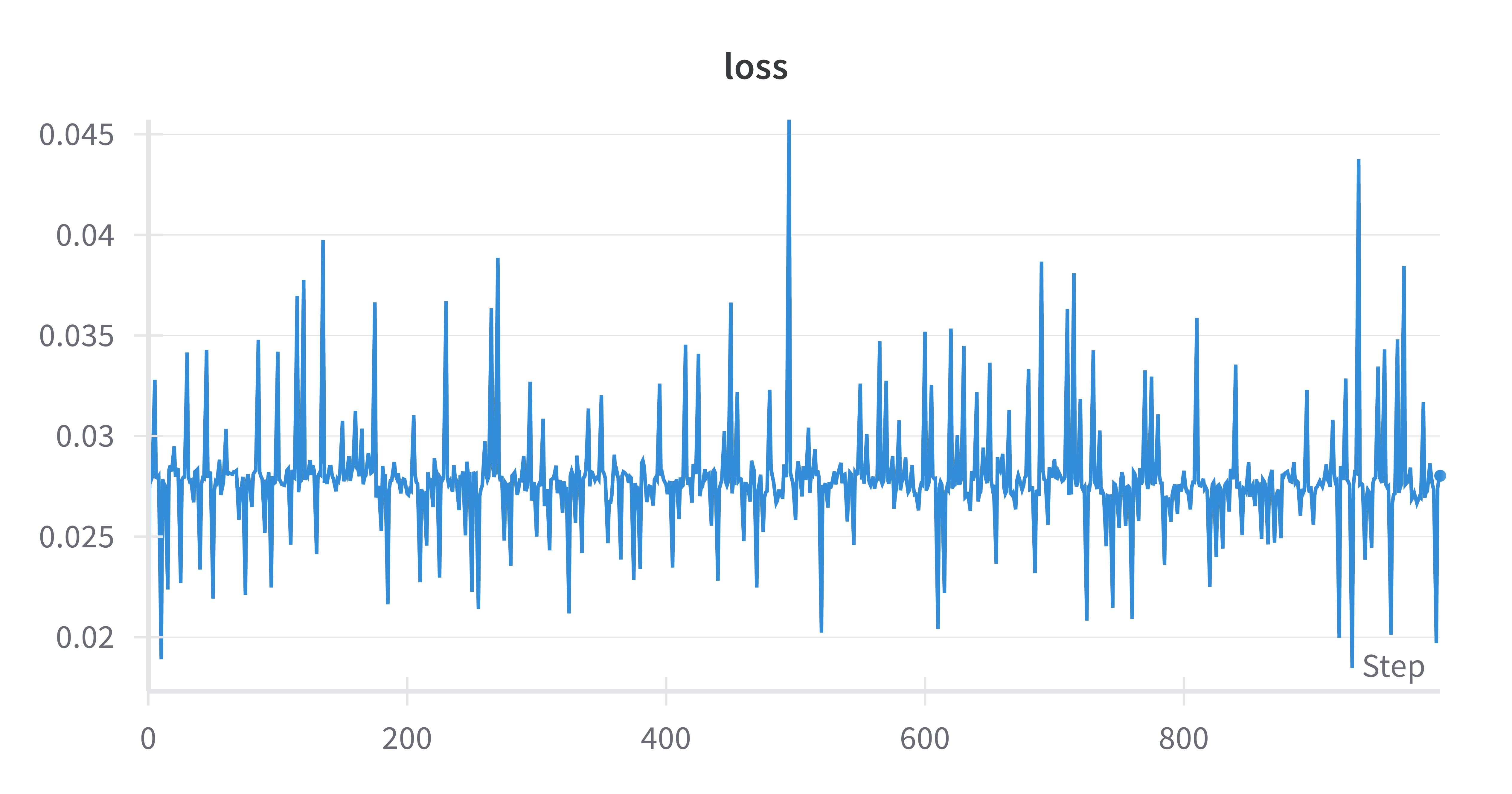} 
    \caption{Loss for the DDIM model after 37,000 steps.}
    \label{fig:ddim_loss}
\end{figure}

\begin{figure}[H]
    \centering
    \includegraphics[width=\columnwidth]{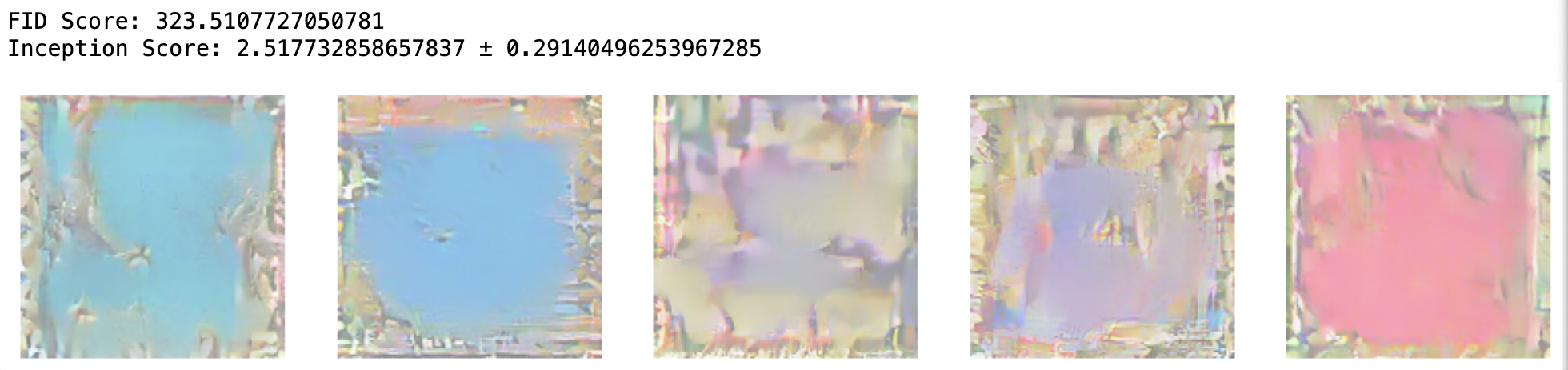} 
    \caption{Results of the DDIM + VAE + CFG model.}
    \label{fig:latent-ddpm_results}
\end{figure}

\begin{figure}[H]
    \centering
    \includegraphics[width=\columnwidth]{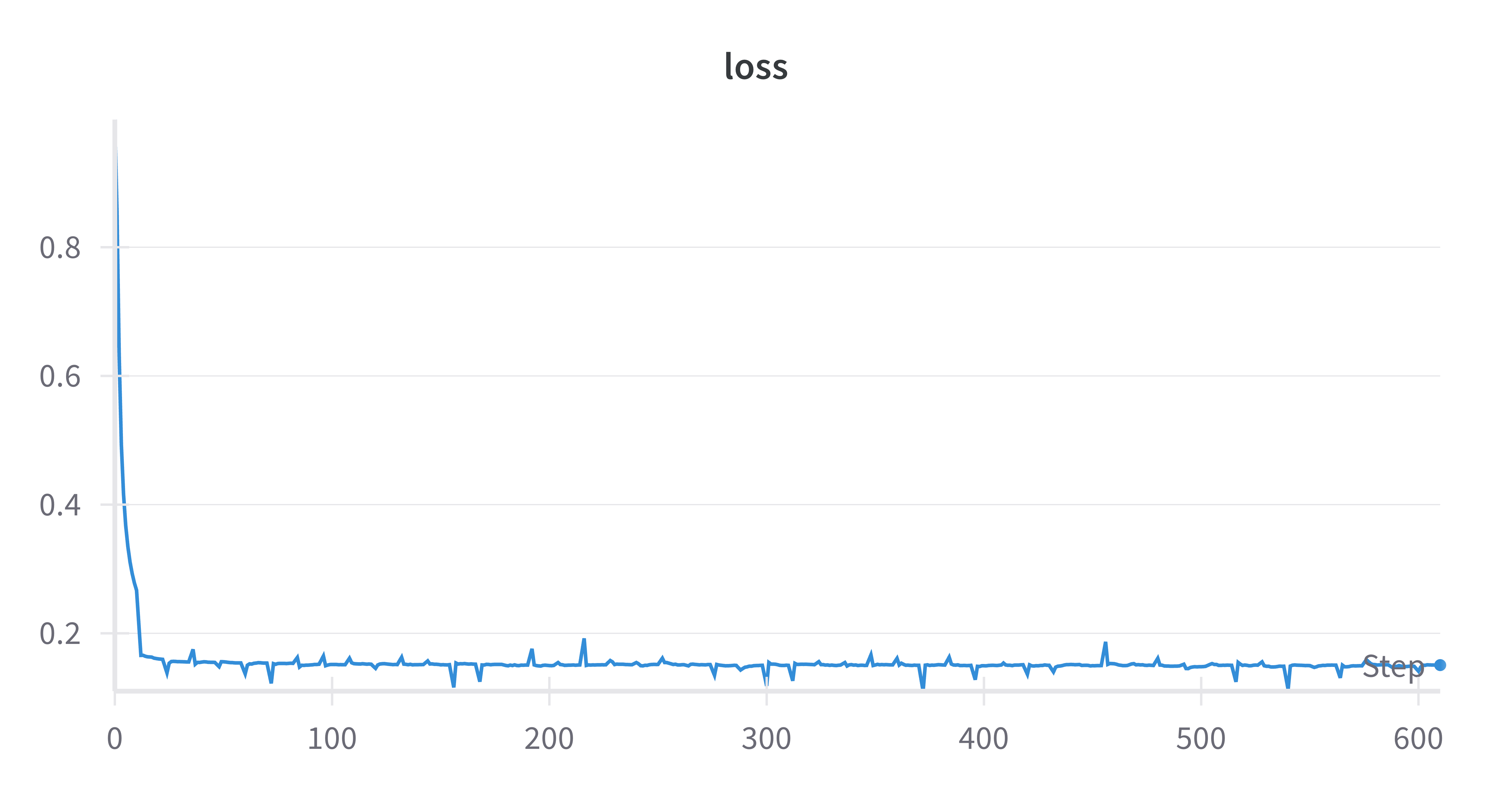} 
    \caption{Loss for the DDIM + VAE + CFG model.}
    \label{fig:latent_loss}
\end{figure}

\subsection{DDIM Efficiency}

The Denoising Diffusion Implicit Models (DDIM) demonstrate a significant improvement in inference efficiency compared to the Denoising Diffusion Probabilistic Models (DDPM). Specifically, DDIM achieves a \textbf{4x speedup} in inference, meaning that for a given model, the inference process with DDIM requires only 25\% of the time needed by DDPM. This improvement is largely attributed to a reduction in the number of inference steps, which decreases from 1000 steps in DDPM to just 250 steps in DDIM. Additionally, DDIM reduces peak memory usage by approximately \textbf{30\%}, further enhancing its practicality and efficiency in resource-constrained environments.

\section{Discussion}

\subsection{Model Efficacy Across Across Experiments}
Our findings indicate that DDIM + CFG consistently outperformed the baseline DDIM with a linear noise schedule. This demonstrates the benefits of deterministic sampling and classifier-free guidance in achieving higher-quality and more diverse image generation. These results align with the performance improvements observed in similar studies on generative models.

\subsection{Comparison with Author Results}

A key aspect of our analysis involves comparing our results with those reported by the original authors of DDPM, DDIM, and Latent Diffusion Models (LDM). While our implementations showed promising trends, they fall short of fully replicating the performance benchmarks established in prior work due to constraints in training duration and computational resources.

\subsubsection{Baseline Comparison: DDPM on CIFAR-10}
The original DDPM paper by Ho et al. \cite{ho2020denoising} achieved an Inception Score (IS) of 9.46 and a Fréchet Inception Distance (FID) of 3.17 on CIFAR-10. Given that our experiment is not complete, we are unable to compare against this baseline currently.

\subsubsection{Efficient Sampling with DDIM}
Song et al.~\cite{song2020denoising} reported significant improvements in inference speed and demonstrated comparable image quality with Denoising Diffusion Implicit Models (DDIM). Specifically, their FID (Fréchet Inception Distance) values on the CIFAR-10 dataset ranged from 4.04 to 13.36, depending on the number of sampling steps used. In contrast, our implementation of DDIM achieved an FID of 47.92. While we successfully reduced the number of sampling steps, this quality gap highlights the critical role of sufficient training epochs in fine-tuning the deterministic sampling process. Additionally, another factor contributing to our underperformance relative to their results is the significant difference in training steps: the authors employed 800,000 steps, whereas our implementation utilized only 61,000 steps. These discrepancies underscore the importance of extended training and careful optimization to achieve state-of-the-art results with DDIM.

\subsubsection{High-Resolution Synthesis with Latent Diffusion Models}
Rombach et al. \cite{rombach2022high} demonstrated the effectiveness of latent diffusion in handling high-resolution datasets like ImageNet, achieving an FID of 27.0 on the full ImageNet dataset. Our results on ImageNet-100, using a pre-trained VAE and CFG, yielded an FID of 323.51. The disparity is largely attributed to:
\begin{itemize}
    \item \textbf{Limited Training Time:} Our models did not train for sufficient epochs to fully adapt to the high-resolution ImageNet-100 dataset.
    \item \textbf{Pre-Trained VAE Dependency:} Our reliance on a pre-trained VAE may have introduced additional constraints, as the VAE was not specifically fine-tuned for our task.
\end{itemize}

\subsubsection{Cosine Noise Scheduler Exploration}
Nichol and Dhariwal \cite{nichol2021improved} reported smoother training dynamics and enhanced sample quality using a cosine noise schedule. This approach mitigates abrupt transitions in noise levels, leading to more stable gradients and improved model convergence. Consequently, the enhanced sample quality highlights the effectiveness of this noise schedule in generating high-fidelity outputs.

\subsection{Challenges with VAE}
The integration of Variational Autoencoders (VAEs) presented notable challenges. Limited training epochs and model capacity constrained performance, highlighting the need for further optimization. These findings emphasize the potential of latent diffusion models, which, with adequate resources, can significantly enhance computational efficiency and output quality.

\subsection{Noise Scheduling}
The cosine noise scheduler exhibited promising results, particularly in enhancing perceptual smoothness during image generation. This observation aligns with the findings of Nichol and Dhariwal \cite{nichol2021improved}, where the cosine schedule demonstrated improved training dynamics and sample quality. However, achieving an optimal balance between diversity and fidelity requires additional tuning and experimentation.

\subsection{Sensitivity and Risks}
The results were particularly sensitive to training duration and dataset scale. Shortened training epochs limited the models' ability to fully adapt, leading to suboptimal noise prediction and image quality. Additionally, the reliance on pre-trained components, such as the VAE, introduced uncertainties regarding their compatibility with our specific tasks.

Potential risks include:
\begin{itemize}
    \item \textbf{Overfitting to Limited Training Data:} Shorter training durations may cause the models to underperform on complex, diverse datasets like ImageNet-100.
    \item \textbf{Dependency on Pre-Trained Models:} The use of pre-trained VAEs could constrain model performance if they are not fine-tuned for the target dataset.
    \item \textbf{Noise Schedule Selection:} Variations in noise scheduling parameters can significantly alter inference performance and require careful tuning.
\end{itemize}

\section{Future Work}

Building upon the challenges and insights highlighted in this study, several directions for future research and development can further optimize the performance and applicability of diffusion models. One significant area of improvement is extending training durations. Prolonged training schedules are essential to enable models to fully converge, refine noise prediction capabilities, and achieve superior image quality metrics such as FID and IS. Leveraging distributed training across multiple GPUs or cloud-based computational resources could make this feasible for larger datasets like ImageNet-100.

Another promising direction involves fine-tuning pre-trained Variational Autoencoders (VAEs) on task-specific datasets. While pre-trained VAEs simplify model integration, they may introduce biases or constraints when not adapted to the target data. Transfer learning approaches, such as retraining the VAE encoder and decoder on datasets like ImageNet-100, can better align latent representations with the diffusion model.

Further investigation is warranted into the sensitivity of classifier-free guidance (CFG) parameters. CFG plays a critical role in balancing fidelity and diversity in generated images, and systematic exploration of its weights across different datasets can provide valuable insights into achieving optimal results. Additionally, alternative noise scheduling strategies, such as exponential or adaptive schedules, should be explored. Noise scheduling profoundly affects the model’s ability to balance global and fine-grained features during training, and novel schedules could dynamically adjust noise addition based on model performance at each timestep.

Scaling models to generate higher-resolution images, such as 256x256 or 512x512, is another critical avenue. This advancement is crucial for applications in industries such as design, gaming, and advertising. Combining improved latent diffusion techniques with larger UNet architectures and hierarchical noise reduction frameworks can make high-resolution synthesis viable. To enhance image realism, experimenting with adaptive loss functions, such as perceptual losses based on pre-trained feature extractors like VGG, could improve the balance between global structure and local details.

Generalization and robustness could also benefit from expanded data augmentation techniques. While current augmentations, such as horizontal flipping, are effective, introducing advanced methods like CutMix, RandAugment, or adversarial perturbations can expose the model to greater data variability, improving its robustness. Additionally, efficient sampling strategies should be explored to reduce the number of sampling steps while maintaining image quality. Hybrid approaches that combine deterministic and stochastic steps or leverage knowledge distillation techniques are potential solutions for reducing inference latency, which is crucial for real-time applications.

Finally, the evaluation of models on diverse datasets beyond CIFAR-10 and ImageNet-100 is imperative. Validating performance on datasets with higher complexity or domain-specific constraints, such as CelebA-HQ, LSUN, or others, can further benchmark the model’s robustness and generalizability. Translating these advancements into real-world applications, such as augmented reality, robotic vision, or creative content generation, is a critical step toward bridging the gap between theoretical progress and practical utility. Collaborating with industry partners to identify use cases and optimize models for deployment constraints, such as edge computing or cloud platforms, will be essential for the broader adoption of diffusion models.

By addressing these areas, future work can advance the state of the art in diffusion models, bridging theoretical advancements with practical utility, and pushing the boundaries of generative AI.

\section{Conclusion}
This project set out to address the computational inefficiencies and scalability challenges inherent in diffusion models for high-quality image generation. The primary goal was to enhance Denoising Diffusion Probabilistic Models (DDPMs) and Denoising Diffusion Implicit Models (DDIMs) to achieve faster inference, improved image quality, and broader applicability across datasets such as CIFAR-10 and ImageNet-100. These challenges are critical to enabling diffusion models to scale effectively for real-world applications in fields like media content creation, gaming, and robotics.

Our findings demonstrate significant progress toward these objectives. The integration of Classifier-Free Guidance (CFG) with DDIM resulted in faster inference and improved image fidelity, addressing the limitations of baseline DDPMs. The use of Variational Autoencoders (VAEs) facilitated efficient latent space compression, enabling training on larger datasets such as ImageNet-100 without compromising computational efficiency. Furthermore, the implementation of a cosine noise scheduler introduced smoother transitions during the denoising process, leading to better sample quality and more efficient training dynamics. These enhancements highlight the potential of combining advanced techniques to overcome the shortcomings of traditional diffusion models.

However, the project also revealed challenges and areas for improvement. The integration of VAEs, while effective, was limited by constraints in training epochs and model capacity, underscoring the need for further optimization. Similarly, the trade-off between image diversity and fidelity when using the cosine noise scheduler indicates that additional tuning is required to fully realize its benefits.

In light of these findings, the project achieved its original goal of demonstrating enhancements to diffusion models and addressing key inefficiencies. While not all challenges were fully resolved, the advancements made in inference speed, sample quality, and computational efficiency provide a strong foundation for future work. These results have significant implications for scaling diffusion models to practical, industry-relevant applications, emphasizing their transformative potential in generative AI.

Future directions include optimizing latent diffusion techniques, refining noise scheduling strategies, and exploring applications across diverse domains. By building on the progress demonstrated in this study, diffusion models can be further advanced to meet the demands of increasingly complex and large-scale generative tasks.

\section*{Authors' Contributions}
The authors contributed to the project as follows:
\begin{itemize}
    \item \textbf{Jaineet Shah:} Implemented the baseline DDPM and the advanced methodologies.
    \item \textbf{Michael Gromis:} Focused on troubleshooting issues with Jaineet, fixing implementation, and created the video and wrote the final report.
    \item \textbf{Rickston Pinto:} Implemented the DDIM improvements and wrote the details to the report.
\end{itemize}

\bibliographystyle{ieeetr}
\bibliography{references}

\end{document}